\documentclass[conference]{IEEEtran}
\IEEEoverridecommandlockouts
\usepackage{cite}
\usepackage{amsmath,amssymb,amsfonts}
\usepackage{algorithmic}
\usepackage{graphicx}
\usepackage{textcomp}
\usepackage{xcolor}
\usepackage{gensymb}
\usepackage{caption,subcaption}
\usepackage{makecell}

\def\BibTeX{{\rm B\kern-.05em{\sc i\kern-.025em b}\kern-.08em
    T\kern-.1667em\lower.7ex\hbox{E}\kern-.125emX}}

\newcommand{\figref}[1]{\figurename~\ref{#1}}
\newcommand{\secref}[1]{Sect.~\ref{#1}}		

\begin{document}

\title{Audience measurement using a top-view camera and oriented trajectories\\
{
\thanks{This work has been developed in the framework of projects TEC2016-75976-R, financed by the Spanish Ministerio de Econom\'{i}a y Competitividad and the European Regional Development Fund (ERDF) and 2014-DI-067 financed by the Catalan government AGAUR.}
}}

\author{\IEEEauthorblockN{Manuel L\'{o}pez-Palma\IEEEauthorrefmark{1},
Javier Gago\IEEEauthorrefmark{1},
Montserrat Corbal\'{a}n\IEEEauthorrefmark{1} and
Josep Ramon Morros\IEEEauthorrefmark{2}
}
\IEEEauthorblockA{\IEEEauthorrefmark{1}Electronic Engineering Dept., Universitat Polit\`{e}cnica de Catalunya, 08222 Terrassa, Spain}
\IEEEauthorblockA{\IEEEauthorrefmark{2}Signal Theory and Communications Dept., Universitat Polit\`{e}cnica de Catalunya, Barcelona, Spain}
}

\maketitle

\begin{abstract}
A crucial aspect for selecting optimal areas for commercial advertising is the probability with which that publicity will be seen. This paper presents a method based on top-view camera measurement, where the probability of viewing is estimated according to the trajectories and movements of the head of the passerby individuals in the area of interest. Using a camera with a depth sensor, the head of the people in the range of view can be detected and modelled. That method allows to determine the orientation of the head which is used to estimate the direction of vision. A tracking by detection algorithm is used to compute the trajectory of each user. The attention given at each advertising spot is estimated based on the trajectories and head orientations of the individuals in the area of interest.

\end{abstract}


\section{Introduction}
\label{sec:intro}

Determining the Focus of Attention (FoA) of individuals in a closed space (for instance, in  commercial areas) -- i.e., %
knowing which are the places and objects that most attract the visual attention of people --
is a problem of wide interest 
whose solution could contribute to 
important 
applications in security, advertisement, commercial distribution, marketing, retail, etc. 

To be useful, a FoA determination model must be robust, non-invasive and adaptable to different environments, often cluttered with many objects. Another important requirement is that the complete setup has to be as cheap as possible to make it competitive in a variety of situations. 
We are proposing a 
method based 
on capturing the scene with a top-view ceiling camera and determining the Focus of Attention of a person by using his/her trajectory and head's direction. While solutions based on eye-tracking could determine more exactly the gaze direction, they are  cumbersome in complex or big scenarios, requiring one high resolution camera at each location at which attention is to be measured. On the other side, the proposed method can estimate the attention at any point of the walls with a single (or few) cameras. Our experiments show that our loss of precision in the gaze direction is not significant in most situations. The top-view configuration presents other advantages, such as being almost immune to occlusions, much less intrusive and less prone to privacy concerns. Moreover, additional information can be obtained, such as the distance of viewing and the relative angle of viewing, thus allowing a richer analysis of the scene.

While detecting the head of 
people 
and its trajectory is more or less straightforward, determining 
a person's head angle 
is more complicated because of the huge variability of 
head types. 
When using 
color cameras the detection of the head angle is often unreliable because of the lack of detail. For this reason we decided to use a camera with a depth sensor. These sensors, instead of capturing information about gray level or color, can measure at each pixel the distance for the camera to the objects in the scene. 
That eases 
the detection of 
heads and also allows to create a 3D model of the head that 
permits to improve the precision of the head orientation estimation. 

Another design decision for 
the architecture 
of our  method 
was not to use deep learning algorithms for head detection/angle estimation/tracking. The reason was to allow the system to run in simple hardware, without the need of a GPU that would increase the cost.

To evaluate the attention given by a viewer to the advertisements  
we use the \textit{Attention/Engagement time (AT)} metric~\cite{Ravnik2013},  
which 
returns the total amount of time the observer is actively looking at 
a sign. AT allows to quantify the degree of attention a given sign has received.
In this paper we will use 
the Focus of Attention~\cite{Palma2018a}, which is a generalization of the concept of Attention Time. 
That measure quantifies the amount of attention a region receives during a period of time. Moreover, it  takes into account other factors such as the distance of the viewer to the target, the speed of movement of the viewer and the angle of vision respect to the trajectory. 
In that way, a more realistic determination of the attention can be provided.

The main contributions of this paper are:
\begin{itemize}
  \item A method based on a top-view depth camera to estimate the attention on 
  any point located on the room walls.
  \item A system for top-view head detection, tracking and estimation of instantaneous planar head angle.
  \item An experimental validation of the estimation of planar head angle using an inertial sensor that gets rid of the need of manual annotation to obtain ground truth.
\end{itemize}

\begin{figure*}[t]
 \centering
    \begin{subfigure}[b]{0.22\textwidth}
     \includegraphics[width=\textwidth]{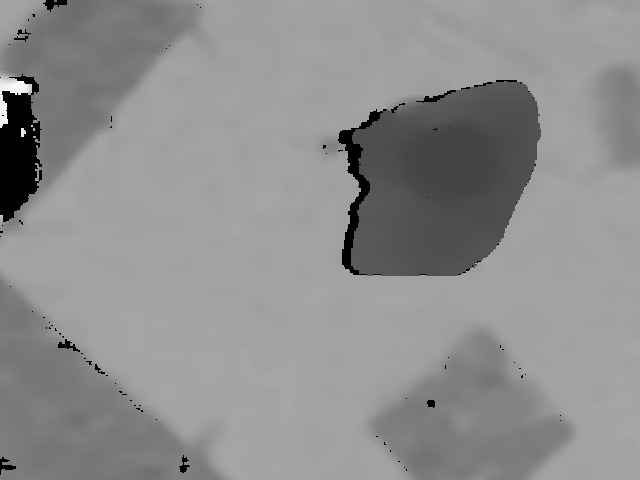}
     \caption{}
    \end{subfigure}
    \begin{subfigure}[b]{0.22\textwidth}     \includegraphics[width=\textwidth]{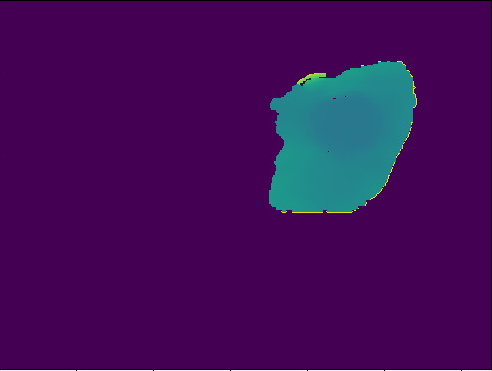}
      \caption{}
    \end{subfigure}
    \begin{subfigure}[b]{0.25\textwidth}
      \includegraphics[width=\textwidth]{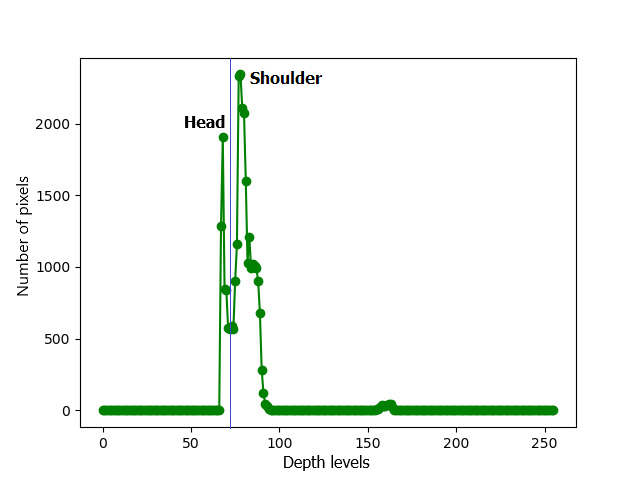}
      \caption{}
    \end{subfigure}
    \begin{subfigure}[b]{0.22\textwidth}
      \includegraphics[width=\textwidth]{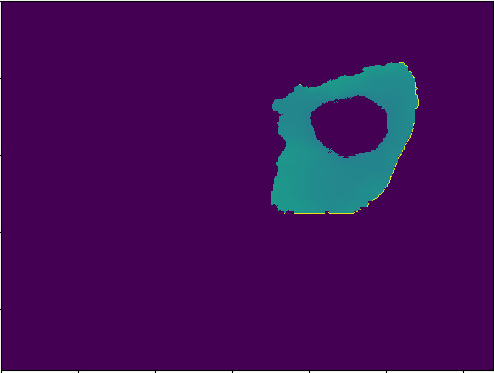}
      \caption{}
    \end{subfigure}
 \caption {a) Depth image b) Body mask c) Body mask histogram d) Head mask 
 }
    \label{fig:detection}
\end{figure*}

This paper is organized as follows: \secref{sec:rel_work} provides a review of the state-of-the-art 
of audience measurement and related technologies. \secref{sec:head_detection} explains the proposed 
 method to detect 
and track a person's head and to determine its various orientations. 
In~\secref{sec:sign_analysis}, some important concepts are formalized. The proposed system is detailed in~\secref{sec:density}. Experimental validation of the proposed system is given in~\secref{sec:exp_res}. Finally, conclusions are drawn in~\secref{sec:concl}.

\begin{figure}[b]
 \centering
      \includegraphics[width=0.55\columnwidth]{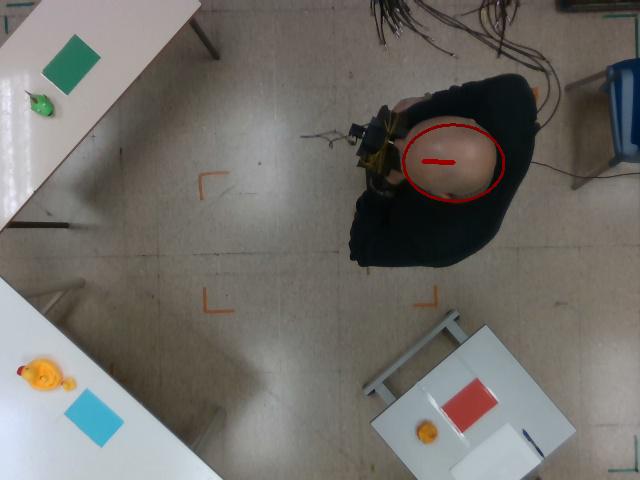}
      \caption{Elipse mask and direction}
    \label{fig:direction}
\end{figure}


\section{Related work}
\label{sec:rel_work}
A crucial aspect affecting the efficiency of advertising is to know 
what are the regions that attract most the Focus of Attention of passerby viewers. 
%
Most methods set their base principle on gaze tracking by using a camera located on the viewed location. For instance,~\cite{Sippl2010} estimates what area of a screen (divided into four quadrants) is looked at by people.
Since the screen exceeds the field of vision, the viewer needs to move his/her head to visualize a quadrant of interest.
This method is based on determining the relative positions of facial features, recognized from a single camera on top of the screen. They assume that the eyes are directed straight ahead. They take into account four movements: up or down, right or left.
%
For the left-right movement they started by focusing on the eyes and the nose tip, but then they found out that the best distinction to take into account 
for a left to right view was the distance from a left to right eye to the nose tip. 
The vertical gaze estimation follows a quite simple approach and can be categorized as a tracking method \cite{Murphy-Chutorian2009}. The basis for the prediction of the vertical gaze is the forehead position. Over a history of 15 frames, the minimum and maximum positions of the forehead are stored. The mean of those two values results in a horizontal line. If the current forehead position is above that line, the vertical gaze prediction will be top; if it is below, it will be bottom. In both cases, the facial features are calculated by an object recognition engine. 
 ~\cite{Clement2015} investigates how the visual prominence over the products in a store affects customer decisions. 
 Frontal cameras are located on or near the product to capture 
 a front view of the person. In these frontal positions, the face and eyes of people can be detected, making possible a fine analysis of the direction of the look. To evaluate the method, the people who participate are required to pass a survey. Then they are captured and analyzed to establish their visual attention and 
 eyes track. Finally, they perform a correlation between visual attention and intended choice.




In 
those frontal positions the face and eyes of people can be detected, making possible a clear-cut analysis of the direction of the look. The placement of the camera determines the applicability of the algorithms for frontal cameras. Located at the ad, the position of the face and eyes of the customer can be detected, making possible a sharp analysis of the direction of the gaze. Its drawbacks are that capturing the face incurs in privacy issues and that they require a camera at each analysis position, which can be affected by occlusions. On the other hand, top-view ceiling cameras are a non-invasive method, can avoid privacy and occlusion problems. A single camera can analyze several spots resulting in cost-effective solutions. The drawbacks are that the face and eyes of the persons are not visible, so the head orientation is used to approximate the gaze direction, and that information such as age and gender can not be determined.

As for the type of sensors, RGB-D sensors take the advantages of the color image that provides appearance information of an object and also of the depth information that is immune to the variations in color and illumination. Many works have used RGB-D cameras \cite{Chen2013, Buys2014, Yamamoto2017, Wu2017, Cai2017}. In~\cite{Chen2013} we provide a review of the use of depth to analyze human activity. RGB-D are popular choices in top-view setups. For instance, in~\cite{Yamamoto2017} customer behaviour is analyzed by locating a RGB-D camera above a shelf. This analysis allows determining which shelf the customer reaches as well as the type of gesture itself. A similar approach is found in~\cite{Wu2017} using a top-view depth camera is used for human posture and activity recognition. That method allows for tracking the users’ positions and orientations and for recognizing their postures and activities (standing, sitting, pointing, and others). In our proposal we use the depth camera in top-view position to measure the focus of attention.



\section{Head detection and tracking}
\label{sec:head_detection}
To detect the head of one or more persons we use the depth information captured from the top-view camera (See \figref{fig:detection} a)). First, background subtraction is performed using the depth image and a model of the background (recorded without any person in the scene). 
That removes any still object
from the scene. A binary mask is generated around the areas showing differences with the background.
After that, small blobs are removed because they represent small objects and not persons. The remaining 
 blobs are potential detections of persons, in particular of their head and shoulders (See \figref{fig:detection} b)). 
To discard objects of a similar height than the one of persons, the depth histograms of the pixels inside each blob are computed. These histograms are compared with a set of reference histograms of the persons' heads, previously computed and stored. We use the correlation between histograms as a similarity measure (HISTCMP\_CORREL in OpenCV\footnote{https://docs.opencv.org/3.4/d6/dc7/group\_\_imgproc\_\_hist.html}).
In this work, we have used five reference histograms, removing  those blobs in which none of the five comparisons is above 0.2.

The remaining detections are assumed to be heads. In that  case, the histogram has two peaks, one for the shoulders region and other corresponding to the head region (See \figref{fig:detection} c)). The head depth levels are determined by computing the histogram minimum between these two peaks and keeping just the pixels with depth levels above the minimum.  
That results in a head mask (See \figref{fig:detection} d)). By fitting an oriented ellipse to this mask we can determine the position (the center of the ellipse) and the orientation of each head (see \figref{fig:direction}) but not its direction.

\begin{figure}[t]
    \centering  \includegraphics[width=0.80\columnwidth]{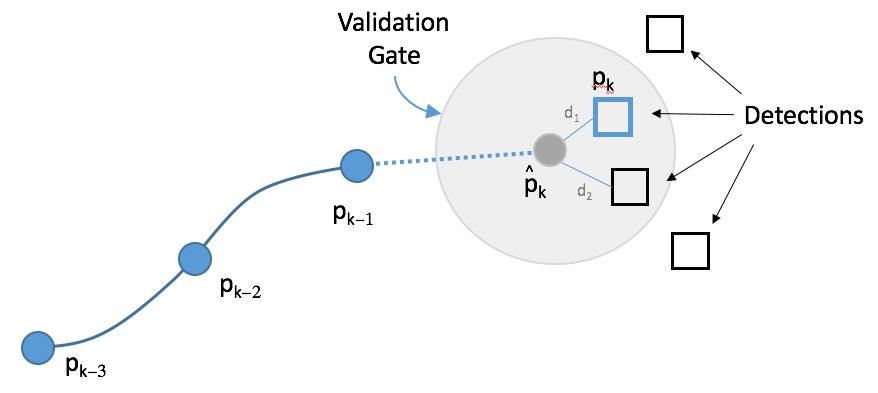}
    \caption {A circular gating function is used to eliminate spurious detections. Nearest Neigbor association decides from the remaining candidates (the ones in the gray circle). In this case, $d_1 < d_2$, so the blue detection is selected.}
    \label{fig:circle}
\end{figure}

To track the heads along the video sequence, a simple association method between tracks and detections is used: for each person we assume a motion model with constant acceleration, where the velocity and acceleration are estimated using the head position in the previous three frames. We use this motion model to predict the position of the person in the next frame. Then, we use a circular gating function~\cite{Geng2017} to discard spurious detections and Nearest Neighbor decision using the euclidean distance to perform the final association (\figref{fig:circle}). The selected head detection at time $t$ receives the same ID as the head at $t-1$. If multiple persons are to be tracked, the Hungarian algorithm could be used to disambiguate multiple matches.


The temporal analysis of the tracking allows also to properly compute the direction of the head (see \figref{fig:direction}): we compute the direction at the initial detection based on the entry point in the camera field and the motion of the head. In the subsequent frames, the change in direction from frame to frame is restricted and no $180 \degree$ turns are allowed from frame to frame. 

\begin{figure}[b]
    \centering  \includegraphics[width=0.80\columnwidth]{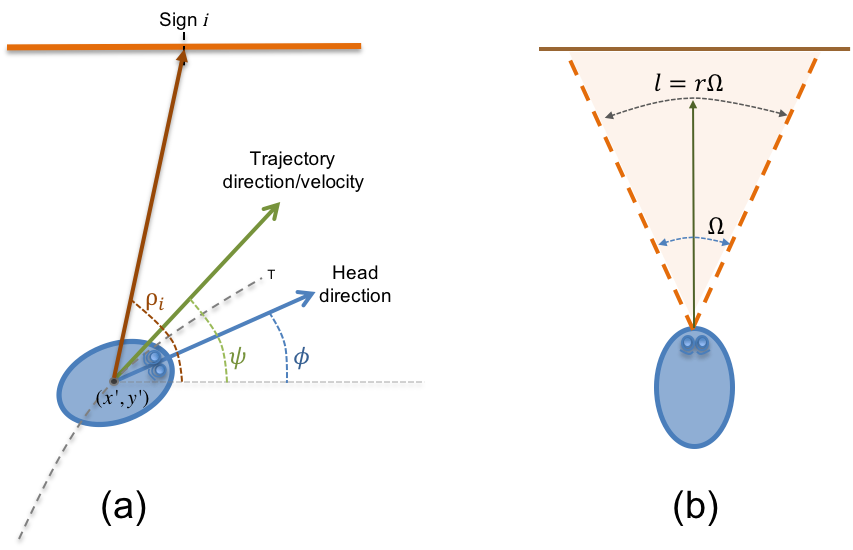}
    \caption {a) State variables b) Angle of vision}
    \label{fig:state_vars}
\end{figure}

\section{Oriented trajectories}
\label{sec:sign_analysis}
The oriented trajectories are computed by determining, at each frame 
captured from a top-view camera, the position $(x',y')$ and head orientation of the user, relative to the room coordinates. 
For simplicity, we assume a rectangular room and the camera to be aligned with the room.

A person inside the room can be parametrised using 
a state vector $\textbf{x}$:
\begin{equation}
\label{eq:state_var}
\textbf{x} = [p', v, \psi, \phi, \theta]
\end{equation}

where $p' = (x',y')$ indicates the position of the person in the room coordinate system, $v$ is the instantaneous velocity of the person, $\psi$ is the direction defined by the person's trajectory, $\phi$ is the planar angle of the head and $\theta$ is the up-down angle of the head (pitch), all in the room coordinate system (see \figref{fig:state_vars} (a)).
In this work we will evaluate the attention to advertisements on the room walls so we will not consider the pitch angle $\theta$.

An oriented trajectory is defined as the temporal sequence of states for all the times instants $k$ a person is in the 
field of view of the camera: $\mathbf{T} = \{\textbf{x}_k\}$.

For each of the objects of interest (signs) we also capture at each time instant the angle that forms the line connecting the center of 
the head and the center of the sign, $\rho_{k}$ (see \figref{fig:state_vars} (a)). The methods described in \secref{sec:head_detection} are used to determine the trajectory and the head angle.

\section{Density of attention and audience measurement}
\label{sec:density}
For audience measurement at a given object (sign on the walls) we will use the aggregation of the Focus of Attention~\cite{Palma2018a} over this object. This measure is a generalization of the purely temporal metric Attention Time (AT) measure~\cite{Ravnik2013}. The AT does compute the amount of time a user (or a set of users) looks directly at a given advertisement. The Focus of Attention introduces ponderative factors such as the distance between the user and the ad, the direction of the gaze of the user and the velocity of the user, assuming that these factors affect the real attention the user pays to the ads. For the sake of completeness we will provide also the temporal metrics AT and IVT defined in \secref{sec:intro}.

Full details about the computation of the FoA 
measure can be found in~\cite{Palma2018a}. For completeness, we will describe here the essential process.




\subsection{Computation of the focus of attention}
To compute the FoA, we determine for each detected person in the scene, the direction of visualization (in our system is given by the direction of the head). Using 
that direction we can determine the  attention cone. The attention cone represents the field of view where it is considered that the user pays attention to what he/she sees. At each time instant $k$, the wall points that lie inside this attention cone receive some degree of attention $A(r)$ that depends on the distance $r$ from the user to the wall. 
That attention can be modeled by:

\begin{equation}
A(r) = C_1 / r
\end{equation}

The constant $C_1$ is the same for all the different persons and can be determined by normalizing the probability maps at the last step of the process. 
That attention is modified by other factors such as the velocity  of  the  person  and  the  relative position of the head with respect to the person’s trajectory. We consider  that  the  degree  of  attention  varies  according  to  the walking speed as:

\begin{equation}
A(v) = \frac{1}{\kappa + v}
\end{equation}

where $\kappa$ is a small regularization constant and $v$ is given by the difference of the positions of the head in successive frames. 
The effect of the angle of visualization ($\phi$) with respect to the trajectory direction ($\psi$) is modeled by a function depending on this angular difference:

\begin{equation}
A(|\psi-\phi|) = 1 + C_2| \psi - \phi|
\end{equation}

The complete instantaneous attention function for a point $\mathbf{p} = (x,y)$ given that the person's
head is located at $\mathbf{p'}=(x',y')$ and oriented along $\phi$ is obtained as a product of all the partial attentions:

\begin{equation}
\label{eq:inst_attention}
A(\mathbf{p}, X) = A(r) \cdot A(|\psi - \phi|) \cdot A(v) 
\end{equation}

The FoA for a given trajectory consists of accumulating the attention function $A$ in \eqref{eq:inst_attention} in the interval $0:k$:

\begin{equation}
\label{eq:att_trajectory}
A_{i}(\mathbf{p}) = \frac{\sum\limits_{k}A^{k}_{i}(x,y, X^k_i)}{\sum\limits_{p}\sum\limits_{k} A_{i}(x,y,X^k_i)} 
\end{equation}

This gives an indication of the normalized attention of an individual at each wall point. This is, the likelihood of each point to be observed by the individual.

To evaluate the attention provided by multiple trajectories (multiple individuals), the individual trajectories will be added and normalized.

\begin{equation}
\label{eq:att_mult_trajectory}
A(\mathbf{p}) = \frac{\sum\limits_{i=1}^{N}A_{i}(p)}{\sum\limits_{i=1}^{N}\sum\limits_{p} A^{i}(p)} 
\end{equation}

\section{Experimental results}
\label{sec:exp_res}
The purpose of this experimental validation section is to demonstrate the ability of the proposed method to determine the relative amount of attention given to different regions of the room. For this, we have captured a set of recordings with a ceiling camera. In the recordings, several individuals walk in predetermined trajectories across the room, while looking at four posters affixed to each wall of the room. The analysis is performed over these posters, trying to determine the degree of attention received by each one of them.

\begin{figure}[b]
    \centering    \includegraphics[width=0.65\columnwidth]{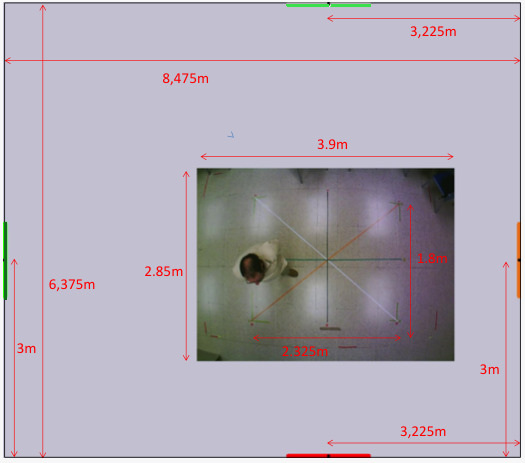}
    \caption {Experimental setup}
    \label{fig:exp_setup_new}
\end{figure}

\figref{fig:exp_setup_new} shows an illustration of the experimental setup. The figure presents a top-view diagram of the room with the location of the four posters (red, orange, green and dark green). The fact that the walls (and posters) are outside the field of view of the ceiling camera doesn't interfere with the proposed algorithm. 

The picture at the center of the figure shows the view from the ceiling camera. The posters are colored A4 paper, glued to the wall at the height of 150 cm from the floor. During the recording, the persons are allowed to walk freely inside the field of view of the depth camera.

Person's trajectories are determined using the method presented in \secref{sec:head_detection} and the 
FoA measures are computed using the equations presented in \secref{sec:density}, in particular 
\eqref{eq:att_mult_trajectory}.

\begin{figure*}[t]
    \centering    
        \begin{subfigure}[b]{0.40\textwidth}
            \includegraphics[width=\textwidth]{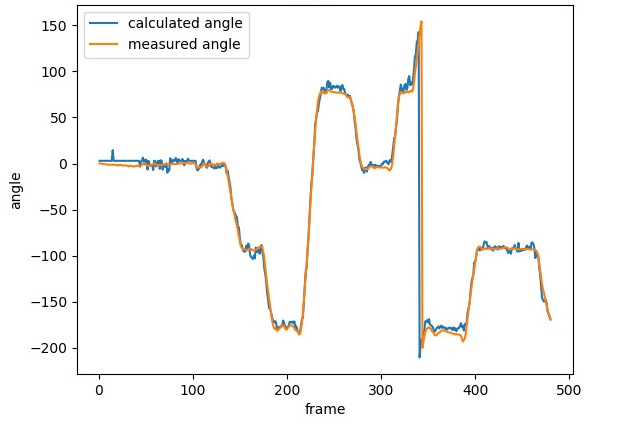}
        \end{subfigure}
        \begin{subfigure}[b]{0.40\textwidth}
            \includegraphics[width=\textwidth]{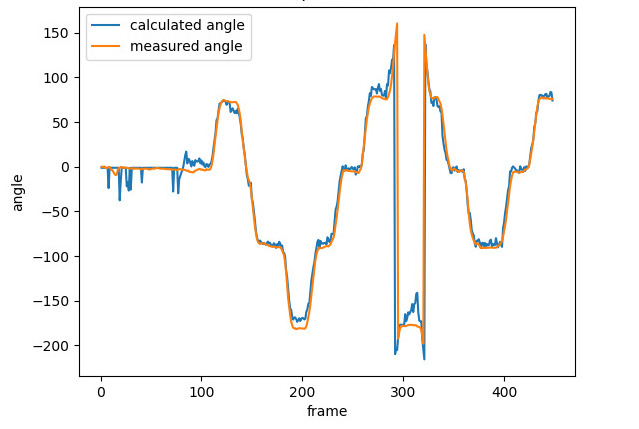}
        \end{subfigure}
    \caption {Results for the head orientation estimation in two different recordings}
    \label{fig:ang_error}
\end{figure*}

To validate the various measures, we have attached a camera and an inertial sensor over the mouth of the various individuals appearing at the recordings. The camera has the same field of vision than the individual bearing it and will be used to validate the attention each add receives from the persons. The inertial sensor (in fact, a high-end Android mobile phone) provides a measure of the angles of the head and will be used to validate the determination of the vision angle (i.e. the head orientation) at each frame. To calibrate this sensor, we asked the persons to enter the scene always from the same side and with the same orientation. 

All the data adquisition (top-view video, front-face video, inertial sensor)
is performed using ROS~\cite{Quigley09}. This type of acquisition allows synchronizing the various data feeds.

In total we have analyzed nine recordings of nine different persons, six men and three women, with different appearances (long hair, bald, and others) and different heights, with a total time of 1108~s. Images are captured at 4 fps.

\textbf{Head detection}: We have analyzed the performance of the head detection algorithm by manually annotating the number of heads visible in all the frames of the test sequences. In the recordings, when 
a person enters the scene, there are a few frames where its head is not completely visible. In this region we obtain a large number of false positives (FP). However, using a post-processing after the tracking step, we can remove all the FP simply by discarding the detections that do not move into the scene. Once the person is fully visible, the number of false positives is zero in our tests. There are some False Negatives (FN) (misdetections), which represent a 3.3\% of the total number of heads in the scene. However, the tracking phase can recover from these misdetections.

\textbf{Head orientation estimation}: To validate the method used to estimate the planar head angle, we compare the results of the algorithm with the measures of the inertial sensor. The metric that has been used is the Mean Absolute Error (MAE) between the angles measured by the inertial sensor and the angles estimated using our algorithm. Overall the test sequences, we obtain a $MAE = 10.69\degree$, which represents an error of $2.97\%$.  \figref{fig:ang_error} shows a graphical comparison between the measured and predicted angles for two recordings. We can see that the method provides a very good estimation of the head orientation.

\textbf{Tracking}: In the analyzed sequences, there is only one visible person at each time instant, and the velocity of the people is not high. In these conditions, the tracking method achieves perfect accuracy. Note that the analysis is most useful in the locations where the persons stay for a given time (for instance, when looking at a shelf or a similar situation) so the conditions of our analysis make sense.

\textbf{Overall algorithm}:
The goal of the algorithm is to determine the relative attentions to different ads or different zones of a room. This would be useful to decide if a product is receiving more attention than another of if people pay more attention to an ad of another company. To validate the full system, we will measure the relative attentions to the four signs on the walls (see~\figref{fig:exp_setup_new}). In order to determine the total attention for a given sign, we accumulate the Attention measure in \eqref{eq:att_mult_trajectory} for all the points of the add normalized by the attention over all the walls. Note that we do not take into account the height information, so the room is considered planar (2D). In fact, we are measuring the attention over a line on the walls (including the ads) and in our approximation, only the width of the ad is relevant. 

The results are summarized in Table~\ref{tab:rel_att}. The first row shows the Attention accumulated over each ad \eqref{eq:att_mult_trajectory}. Higher values mean more attention paid to the add. The following procedure is used to validate these measures: A frontal camera located over the mouth of the individuals records the scene. In these videos (synchronized with the top-view camera) we detect the ads in the walls. Detection is straightforward using color segmentation because each ad has a distinctive color. With this, we create a ground truth as the time that the ad is visible for each user (by adding the number of frames in which the ad is visible). As this measure is not directly comparable to the Attention measure, we normalize it by dividing by the value of the most viewed ad. The results are given in the second row of the table I as the percentage of time the ad is viewed respectively to the most viewed one. 

\begin{table}[h]
\caption{Relative attentions}
\begin{center}
\begin{tabular}{|c|c|c|c|c|}
\hline
&\multicolumn{4}{|c|}{\textbf{Observed Ads}} \\
\cline{2-5} 
\textbf{Method} & \textbf{\textit{Orange}}& \textbf{\textit{green}}& \textbf{\textit{red}}& \textbf{\textit{dark green}} \\
\hline
\textbf{Accumulated Attention} & \textbf{\textit{0.299}}& \textbf{\textit{0.124}}& \textbf{\textit{0.023}}& \textbf{\textit{0,001}} \\
\hline
\makecell{\textbf{Ground truth} \\(relative to max.)} & \textbf{\textit{100\%}}& \textbf{\textit{53,5\%}}& \textbf{\textit{15,4\%}}& \textbf{\textit{0,02\%}} \\
\hline
\makecell{\textbf{Attention}\\ (relative to max.)} & \textbf{\textit{100\%}}& \textbf{\textit{41,6\%}}& \textbf{\textit{7,7\%}}& \textbf{\textit{0,19\%}} \\
\hline
\end{tabular}
\label{tab:rel_att}
\end{center}
\end{table}

We repeat the same process over the Attention measure that is obtained through the trajectory and the results are presented in the third row. Here we have also normalized 
it by dividing by the value of the most viewed ad. We see that the relative order is preserved for both measures: the most viewed is the orange ad, then the green, the red and finally, the dark green. The actual figures of the relative attention are similar enough, even if the Attention method takes into account more factors (distance being one of them) than the time of viewing alone.

\section{Conclusions}
\label{sec:concl}
We have presented a new Top-View Depth Camera Method to estimate the Focus of Attention given by a person to any point located within a room walls.  
The person's head is detected by means of the depth information captured from a top-view camera and its orientation is estimated.
A simple association method between tracks and detections has been applied to track head positions when the person moves around the room.  
We also instantiate an experimental  method to validate the results based on locating an inertial sensor and a frontal camera on the user. Such a method has the advantage of returning ground truth values without the need of manually annotating sequences.
Experimental results prove that our algorithm is capable of determining the relative attention given by a viewer to different regions of the walls.

\bibliographystyle{IEEEtran}
\bibliography{references.bib}   

\end{document}